\documentclass[letterpaper, 10 pt, conference]{ieeeconf}  % Comment this line out if you need a4paper

\IEEEoverridecommandlockouts                              % This command is only needed if 
\usepackage{graphics} % for pdf, bitmapped graphics files
\usepackage{amsmath} % assumes amsmath package installed
\usepackage{amssymb}  % assumes amsmath package installed
\usepackage{makecell}
\usepackage{hhline}
\usepackage{graphicx}
\usepackage[]{todonotes} %disable
\usepackage{siunitx}
\usepackage{booktabs}
\usepackage{tabularx}
\usepackage{multirow}
\usepackage{mwe}

\newcommand{\includeimage}[4]{
	\begin{figure}
		\centering
		\includegraphics[#1]{figures/#2}
    	\caption{#3}
    	\label{#4}
	\end{figure}
}

\newcommand{\maketable}[3]{
	\begin{table}
    	\centering
    	\caption{#2}
		{#1}
    	\label{#3}
	\end{table}
}

\newcommand{\maketabletwocolumns}[3]{
	\begin{table*}
		\centering
		\caption{#2}
		{#1}
		\label{#3}
	\end{table*}
}

\title{\LARGE \bf
InEKFormer: A Hybrid State Estimator for Humanoid Robots
}

\author{Lasse Hohmeyer$^{1}$, Mihaela Popescu$^{1}$, Ivan Bergonzani$^{2}$, Dennis Mronga$^{2}$ and Frank Kirchner$^{1,2}$% <-this % stops a space
\thanks{$^{1}$University of Bremen, 28359 Bremen, Germany.}%
\thanks{$^{2}$Robotics Innovation Center, German Research Center for
       Artificial Intelligence (DFKI GmbH), 28359 Bremen, Germany.}%
}

\begin{document}

\maketitle
\thispagestyle{empty}
\pagestyle{empty}

%%%%%%%%%%%%%%%%%%%%%%%%%%%%%%%%%%%%%%%%%%%%%%%%%%%%%%%%%%%%%%%%%%%%%%%%%%%%%%%%
\begin{abstract}

Humanoid robots have great potential for a wide range of applications, including industrial and domestic use, healthcare, and search and rescue missions. However, bipedal locomotion in different environments is still a challenge when it comes to performing stable and dynamic movements. This is where state estimation plays a crucial role, providing fast and accurate feedback of the robot's floating base state to the motion controller. Although classical state estimation methods such as Kalman filters are widely used in robotics, they require expert knowledge to fine-tune the noise parameters. Due to recent advances in the field of machine learning, deep learning methods are increasingly used for state estimation tasks. In this work, we propose the InEKFormer, a novel hybrid state estimation method that incorporates an invariant extended Kalman filter (InEKF) and a Transformer network. We compare our method with the InEKF and the KalmanNet approaches on datasets obtained from the humanoid robot RH5. The results indicate the potential of Transformers in humanoid state estimation, but also highlight the need for robust autoregressive training in these high-dimensional problems.

\end{abstract}

%%%%%%%%%%%%%%%%%%%%%%%%%%%%%%%%%%%%%%%%%%%%%%%%%%%%%%%%%%%%%%%%%%%%%%%%%%%%%%%%

\section{INTRODUCTION}

Humanoid robots are increasingly leaving their controlled lab environments and are used in real-world applications, requiring safety, robustness, and stability when performing locomotion on different terrains. 
State estimation plays a crucial role in achieving stable and robust locomotion of humanoids. 
The overall performance critically depends on fast and accurate computations of the robot's floating base position, orientation and velocity.
These quantities are typically estimated from noisy and incomplete measurements of the feet contacts, leg kinematics, inertial measurement unit (IMU) and, potentially, visual sensors.  
A great deal of research effort has been invested in the development of algorithms for state estimation, which can be broadly categorized into model-based approaches~\cite{hartley2020}, data-driven approaches~\cite{Geneva2022, Goel2024}, as well as hybrid methods, which combine the former two~\cite{revach2022a}. 
The vast majority of the hybrid methods combine Kalman filters with data-driven approaches, such as deep neural networks. The latest research results indicate that these methods are increasingly outperforming both model-based and purely data-driven methods~\cite{Jin2021, Bai2023,Feng2023}. 

On technical level, most of the hybrid algorithms use a data-driven approach in parallel or succession to a model-based state estimator, such as the Kalman filter. Less commonly, the two approaches are interconnected, as in the case of KalmanNet~\cite{revach2022a}. Addressing this gap, this paper explores the full integration of the InEKF~\cite{hartley2020} with a Transformer network~\cite{Vaswani2017}, as shown in Fig.~\ref{fig:hybrid_state_est}. Specifically, the Transformer is used to predict the Kalman gain to better account for model mismatches than a purely model-based approach. For offline training, we created and published a new dataset consisting of motion capture data and proprioceptive measurements of the RH5 humanoid robot~\cite{esser2021, Boukheddimi2022} performing five motion types: walking, squatting, turning, hip movement and single-leg balancing. We evaluate our approach on the task of estimating the floating base state of RH5, comparing it with the InEKF and KalmanNet~\cite{revach2022a} as baselines. 

\includeimage{width=\linewidth}{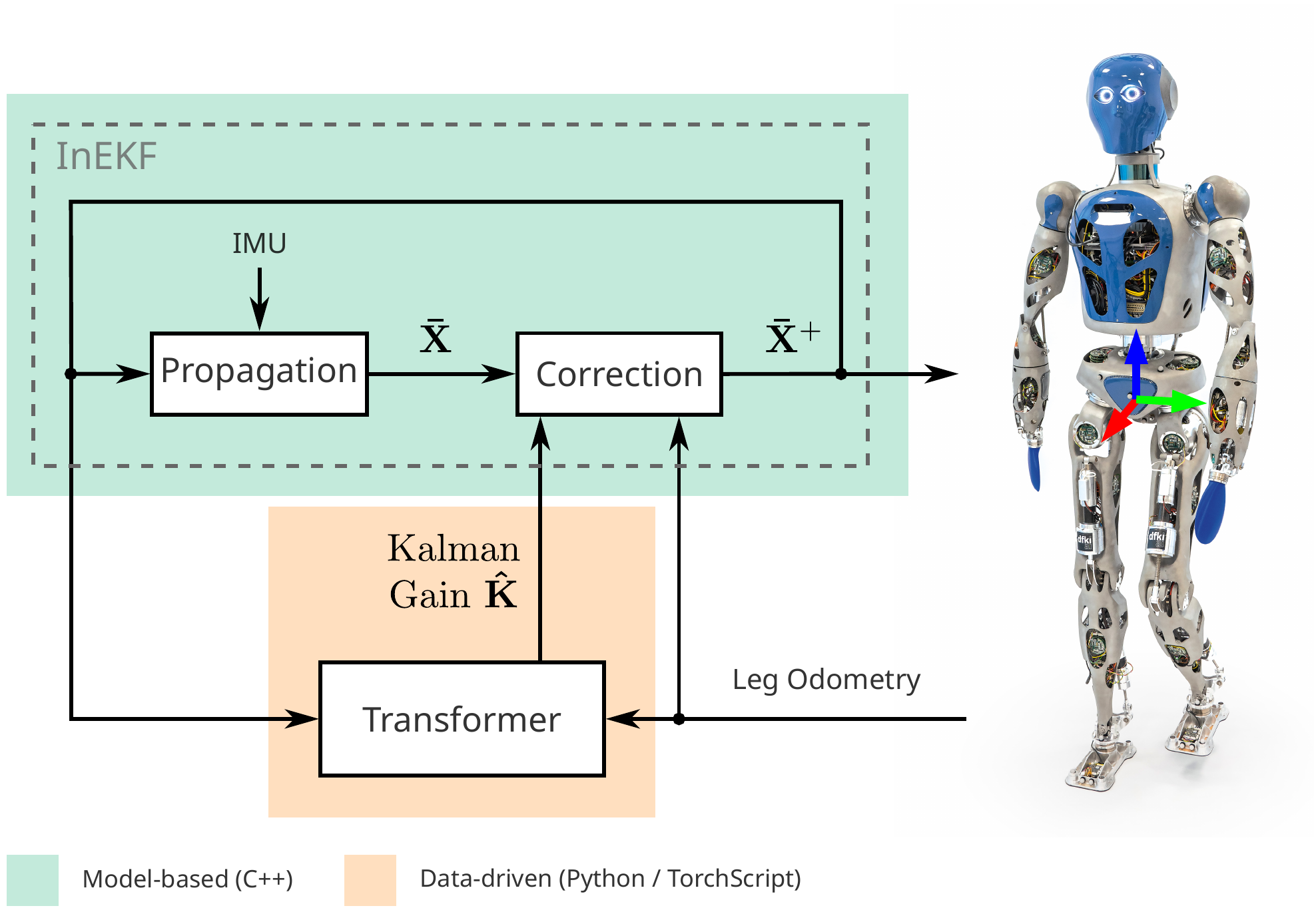}{Hybrid state estimation approach consisting of a model-based InEKF and a data-driven Transformer network for the RH5 humanoid robot.}{fig:hybrid_state_est}

In summary, the contributions of this work are as follows: 
\begin{itemize}
\item A novel hybrid state estimation approach called the InEKFormer, which combines the InEKF with a Transformer network.
\item A humanoid dataset for offline training of the Transformer network with ground truth data from a motion capture system.
\item An extensive comparison of the novel InEKFormer approach with InEKF and KalmanNet.
\end{itemize}

This paper is structured as follows. Section~\ref{sec:related_work} presents the current state of the art and provides the context for our approach. Section~\ref{sec:methodology} describes the methodology. Section~\ref{sec:evaluation} presents the experimental results, and Section~\ref{sec:conclusion} draws the conclusions and points to potential future work. 

\section{RELATED WORK}
\label{sec:related_work}

\subsection{Model-based State Estimation}

The standard approach to perform state estimation for legged robots relies on analytical information of the system model, such as kinematics, dynamics, and noise properties. For instance, \cite{Bloesch2012} introduced an Extended Kalman Filter (EKF) for proprioceptive state estimation of a quadruped robot with point feet, fusing IMU and leg kinematics measurements. This work has been further developed by \cite{Rotella2014} into a quaternion-based EKF for a humanoid robot with flat-foot rotational constraints. However, the performance of the EKF is limited by linearisation errors, which can cause the filter to become inaccurate and diverge.

To address this issue, the InEKF was proposed by \cite{hartley2020} for proprioceptive state estimation on the bipedal robot Cassie. InEKF exploits the Lie group symmetries of the robot state to improve convergence, as linearisation is independent of the current state estimate. However, adding state elements such as IMU accelerometer and gyroscope biases causes the filter to lose its theoretical properties and become "imperfect".

\subsection{Learning-based State Estimation}

Recent advances in deep neural networks, such as recurrent neural networks (RNNs) \cite{Goodfellow2016} specialised in processing sequential data, and Transformer models \cite{Vaswani2017} with attention mechanisms, suggest that machine learning methods could be used as an alternative to model-based state estimation.

The work in \cite{Geneva2022} uses a Transformer network to model physical systems by leveraging the Koopman operator to handle nonlinear systems through a globally linear representation. However, this approach has only been applied to low-dimensional problems, such as the 3D Lorenz attractor. Furthermore, \cite{Goel2024} lays the theoretical foundations for proving that a Transformer can approximate the Kalman filter in a strong sense, though it has not yet been applied in practical scenarios. In general, a disadvantage of deep neural networks is that they require large amounts of training data, which can be difficult to obtain in robotics scenarios. Additionally, as models scale up with state space size, greater computational effort is required for high-frequency state estimation.

\subsection{Hybrid State Estimation}

Recent review papers \cite{Jin2021, Bai2023, Feng2023} suggest a trend towards combining model-based and data-driven methods into hybrid approaches for state estimation. The aim is to exploit the benefits of both techniques by using the known system dynamics from model-based approaches and compensating for unmodelled parameters and noise using data-driven methods.

Hybrid approaches fall into two categories. The first category involves combining model-based and data-driven algorithms \textit{externally} in parallel or in succession, while keeping the internal structure of each algorithm unchanged. In legged robots, hybrid state estimation is typically based on the InEKF model-based estimator due to its superior robustness and convergence properties. The work in \cite{buchanan2022} addresses proprioceptive state estimation by fusing learned inertial odometry with leg kinematics measurements to improve filter performance on slippery and compressible terrains. Furthermore, \cite{lin2022} enhances the InEKF proprioceptive estimator by learning leg contact events on different terrains. In \cite{yang2024}, neural networks are used to learn the weighted mean leg odometry from foot forces and leg kinematics of a hydraulic quadruped robot, which is used as measurement in the InEKF. Including exteroceptive sensing, \cite{schperberg2024} employs a Transformer autoencoder to extract semantic information and robot height from depth images, and then refines the proprioceptive estimation through a Gated Recurrent Unit (GRU) network. Overall, these methods address externally coupled hybrid approaches, either by providing learned input to the filter or by improving its output. However, their application in legged robotics has been limited to quadruped robots and does not explore the use case of humanoids.

The second category refers to \textit{internally} combined hybrid methods, where the data-driven models take over internal functionalities of the model-based algorithms. In KalmanNet \cite{revach2022a}, an RNN is used to obtain the Kalman gain of the EKF to compensate for nonlinearities and model mismatch. Several extensions have been proposed, including the unsupervised KalmanNet \cite{revach2022b} which uses EKF predictions for training, the uncertainty-aware KalmanNet \cite{klein2022} to obtain the state covariance of fully observable systems, and the Latent-KalmanNet \cite{buchnik2024} to improve interpretability. Moreover, \cite{shen2024} proposes the KalmanFormer, which uses a Transformer model instead of RNNs to learn the Kalman gain in an EKF. However, these filters have only been applied to canonical examples with a reduced state space, such as the simple pendulum with two states, the 3D Lorenz attractor, and a planar robot with four states. Moreover, model-based approaches often rely on standard Kalman filters, which have been outperformed by the InEKF in terms of state estimation performance. 

In order to address this research gap, we developed a new hybrid state estimation method that is internally coupled and employs the state-of-the-art InEKF model-based approach \cite{hartley2020}, as well as a powerful Transformer network \cite{Vaswani2017}. Furthermore, we applied our method to a full-size humanoid robot, which has not yet been showcased in literature.

\section{METHODOLOGY}
\label{sec:methodology}

The InEKFormer is a hybrid state estimation algorithm which internally combines the model-based InEKF with a machine learning approach. Unlike the KalmanNet \cite{revach2022a}, a Transformer model is used to estimate the Kalman gain due to its superior contextual interpretation and scalability to larger problem sizes compared to RNNs. Moreover, the state propagation and correction steps, along with the state matrix computation on Lie groups, are adopted from the InEKF \cite{hartley2020}. 

\subsection{Invariant EKF}

The state propagation step of the InEKF involves the state transition function $f_{\mathbf{u}}$, which propagates the current state estimate $\mathbf{\bar{X}}^+_{t-1}$ to the next time step based on the IMU measurements $\mathbf{u}$ and a strapdown IMU model:
\begin{align}
    \frac{\mathrm{d}\mathbf{\bar{X}}_t}{\mathrm{d}t} = f_{\mathbf{u},t}(\mathbf{\bar{X}}^+_{t-1})
\end{align}
On the other hand, the state correction step calculates the corrected state $\mathbf{\bar{X}}^+_t$ using the Lie exponential from the propagated state $\mathbf{\bar{X}}_t$:
\begin{align}
    \mathbf{\bar{X}}^+_t = \exp(\mathbf{K}_t \boldsymbol{\Pi} \mathbf{\bar{X}}_t \mathbf{Y}_t) \mathbf{\bar{X}}_t
    \label{eq:correction}
\end{align}
\noindent where $\mathbf{K}_t$ denotes the Kalman gain matrix, $\mathbf{Y}_t$ denotes the observation vector resulting from leg odometry and $\boldsymbol{\Pi}$ is a selection matrix that selects the first three entries from the innovation vector $\mathbf{\bar{X}}_t \mathbf{Y}_t$.

As the hybrid approach does not perform covariance estimation, the equations for propagating and correcting the state covariance matrix $\mathbf{P}$ have been omitted for conciseness. These equations can be found in \cite{hartley2020}.

\subsection{Hybrid State Estimation Algorithm}

Typically, the Kalman gain is calculated using uncertainty information about past state estimates, incorporating handcrafted noise covariance matrices. However, the InEKFormer circumvents this by implicitly extracting the necessary information from a history of state and observation differences. This makes the method suitable for any robotic platform that relies on state estimation in the presence of model mismatch and sensor noise, such as legged, wheeled, and aerial robots. The high-level algorithm architecture is depicted in Fig. \ref{fig:hinekf}.

\includeimage{width=\linewidth}{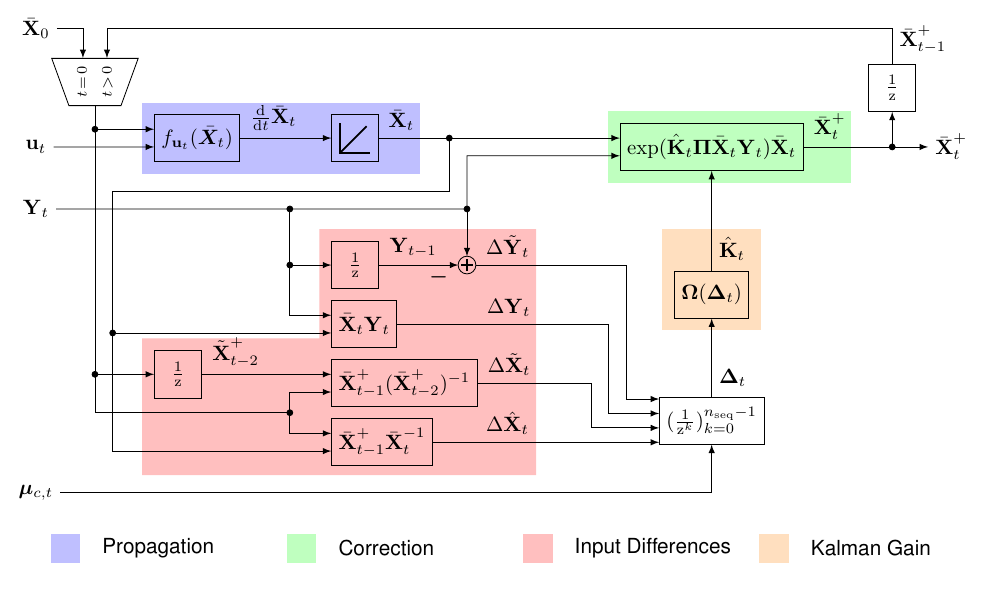}{Block diagram of the high-level InEKFormer architecture, where $\frac{1}{\mathrm{z}^k}$ denotes $k$ unit delays.}{fig:hinekf}

In this work, InEKFormer is applied to the state estimation of the RH5 full-size humanoid robot, which exhibits alternating foot contact states during locomotion. While the model-based InEKF takes into account by adding and removing contacts in the state matrix, the InEKFormer permanently augments the state with left and right foot contacts, as the Transformer requires a fixed input token dimension. The resulting SE$_4$(3) Lie group state matrix $\bf{\mathbf{X}}$~is:
\begin{equation}
\renewcommand{\arraystretch}{1.5}
^\mathrm{W}_{\mathrm{B}}\mathbf{X} = \left(\begin{array}{c|cccc}
    ^\mathrm{W}_{\mathrm{B}}\mathbf{R} & ^\mathrm{W}_{}\mathbf{v}^{}_{\mathrm{B}} & ^\mathrm{W}_{}\mathbf{p}^{}_{\mathrm{B}} & ^\mathrm{W}_{}\mathbf{c}^{}_{\mathrm{L}} & ^\mathrm{W}_{}\mathbf{c}^{}_{\mathrm{R}}\\
    \hline
    \mathbf{0}_{4\times3} & \multicolumn{4}{c}{\mathbf{I}_{4\times4}}\\
\end{array}\right)
\end{equation}
\noindent where $^\mathrm{W}_{\mathrm{B}}\mathbf{R}$, $^\mathrm{W}_{}\mathbf{v}^{}_{\mathrm{B}}$ and $^\mathrm{W}_{}\mathbf{p}^{}_{\mathrm{B}}$ denote the rotation, velocity and position of the robot floating base frame $\{\mathrm{B}\}$ in world coordinate frame $\{\mathrm{W}\}$, and $^\mathrm{W}_{}\mathbf{c}^{}_{\mathrm{L}}$ and  $^\mathrm{W}_{}\mathbf{c}^{}_{\mathrm{R}}$ denote the position of the left and right foot contact frames $\{\mathrm{F}_{\mathrm{L}}\}$ and $\{\mathrm{F}_{\mathrm{R}}\}$ in world coordinates, respectively. The robot coordinate frames are depicted in Fig. \ref{fig:rh5_frames_and_sensors} and explained in detail in Section~\ref{sec:robot_dataset}.

The right-invariant observation vectors $^{\mathrm{B}}_{}\mathbf{Y}_{\mathrm{L}}$ and $^{\mathrm{B}}_{}\mathbf{Y}_{\mathrm{R}}$ corresponding to the left and right foot are:
\begin{equation}
^{\mathrm{B}}_{}\mathbf{Y}_i = [^{\mathrm{B}}_{}\mathbf{h}^{}_i(\boldsymbol\alpha)^{\mathrm{T}}_{}\ 0 \ 1 -\!\!1 \ 0 ]^{\mathrm{T}}_{}, \ i \in \{\mathrm{L}, \mathrm{R}\}
\end{equation}
\noindent where $^{\mathrm{B}}_{}\mathbf{h}^{}_i$ denotes the forward kinematics, i.e., the position of contact $i$ in floating base coordinates $\{\mathrm{B}\}$ given the joint angles $\boldsymbol\alpha$. Since the RH5 robot has flat feet without articulated toes, the contacts are modeled as point contacts. In order to account for the alternating contact states within the fixed-size state matrix, contact information $\boldsymbol\mu_c = [\mu^{}_{\mathrm{L}} \ \mu^{}_{\mathrm{R}}]^{\mathrm{T}}$ is provided to the gain estimator model at each time step. This is obtained by thresholding the contact force in the $z$-direction using the logistic sigmoid function as follows:
\begin{equation}
\mu_{i} = 1 - (1 + \mathrm{e}^{-(F_{T} + F_{\mathrm{z,i}})})^{-1}, \ i \in \{\mathrm{L}, \mathrm{R}\}
\end{equation}
\noindent where $F_{\mathrm{z,i}}$ denotes the contact force in local $z$-direction at foot $i$ and $F_T = 50$ Nm, a value that has been determined experimentally. The contact states in $\boldsymbol\mu_{c}$ are scalar values between 0 and 1 for each foot, where 0 indicates no contact and 1 indicates fully stable contact. This non-binary, continuous contact state format allows for more sophisticated contact estimators in future, such as probabilistic ones.

To obtain the necessary statistical information about the measurement and process noise, as well as the contact states, the calculation of the four input features from KalmanNet is adapted to the InEKF Lie group structure. The five resulting input features for the gain estimator model are:

\begin{itemize}
    \item[\textit{F1}] Observation difference $\Delta\tilde{\mathbf{Y}} = \mathbf{Y}_t - \mathbf{Y}_{t-1}$.

    \item[\textit{F2}] Innovation difference $\Delta\mathbf{Y} = \bar{\mathbf{X}}_t \mathbf{Y}_t$.

    \item[\textit{F3}] Forward evolution difference $\Delta\tilde{\mathbf{X}} = \bar{\mathbf{X}}^+_{t-1} (\bar{\mathbf{X}}^+_{t-2})^{-1}$.

    \item[\textit{F4}] Forward update difference $\Delta\hat{\mathbf{X}} = \bar{\mathbf{X}}^+_{t-1} \bar{\mathbf{X}}^{-1}_t$.

    \item[\textit{F5}] Contact states $\boldsymbol\mu_c = [\mu^{}_{\mathrm{L}} \ \mu^{}_{\mathrm{R}}]^{\mathrm{T}}$.
\end{itemize}

The features \textit{F1} - \textit{F4} are calculated in world coordinates, which makes them right-invariant differences, since the robot state is also expressed in world coordinates. Features \textit{F1} and \textit{F2} are calculated for each foot and concatenated into one vector before being fed into the gain estimator model.

\subsection{Gain Estimator Model}

The gain estimator model uses a Transformer network with an encoder-decoder architecture and scaled dot-product attention to predict the Kalman gain $\mathbf{K}$ from a sequence of input features \textit{F1} - \textit{F5}. The detailed architecture of the gain estimator model is depicted in Fig. \ref{fig:omega}.

\includeimage{width=\linewidth}{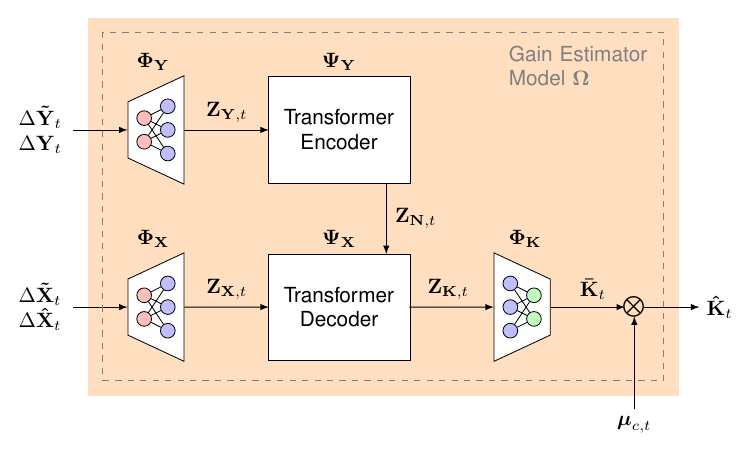}{InEKFormer gain estimator model architecture.}{fig:omega}

Upstream of the actual gain estimator model, the statistical input features \textit{F1} - \textit{F4} are scaled by a robust scaler onto a certain quantile range, which is a hyperparameter to be optimized. The features \textit{F1} and \textit{F2}, which encode statistical information about the observations, are concatenated into a 12-dimensional vector. This vector is then embedded using a learned fully-connected embedding $\boldsymbol\Phi_{\mathbf{Y}}$ and fed into the Transformer encoder $\boldsymbol\Psi_{\mathbf{Y}}$, resulting in a latent representation of the observation noise $\mathbf{Z}_{\mathbf{N}}$. This resembles the InEKF computation of the observation covariance matrix $\mathbf{N}$ from~\cite{hartley2020}.

Conversely, features \textit{F3} and \textit{F4}, which encode statistical information about system noise, are flattened, concatenated to a 42 dimensional vector, embedded by another fully connected embedding $\boldsymbol\Phi_{\mathbf{X}}$ and fed into the Transformer decoder $\Psi_{\mathbf{X}}$ to predict the latent representation of estimation uncertainty $\mathbf{Z}_{\mathbf{K}}$. This, in turn, resembles the covariance propagation step of the model-based InEKF. Finally, the preliminary Kalman gain matrix $\mathbf{\bar{K}} \triangleq \begin{bmatrix} \mathbf{\bar{K}}_{\mathrm{L}} & \mathbf{\bar{K}}_{\mathrm{R}}\end{bmatrix}$ is computed by a fully connected decoder $\boldsymbol\Phi_{\mathbf{K}}$. 

In order to compute the final Kalman gain estimate $\mathbf{\hat{K}}$, the contact information must be taken into account. In principle, it is possible to feed \textit{F5} alongside the other input features into the Transformer model, but this would make the size of the Transformer model dependent on the dimensionality of the contact state representation. To avoid this, the preliminary Kalman gain matrix $\mathbf{\bar{K}}$ is post-processed. This involves multiplying the columns originating from the left and right foot contacts by their respective contact states $\boldsymbol\mu_{c}$  to obtain the final Kalman gain estimate $\mathbf{\hat{K}}$ as follows:
\begin{equation}
	\mathbf{\hat{K}} = \begin{bmatrix} \mu_{\mathrm{L}}\mathbf{\bar{K}}_{\mathrm{L}} & \mu_{ \mathrm{R}}\mathbf{\bar{K}}_{\mathrm{R}}\end{bmatrix}
\end{equation}

\section{EVALUATION}
\label{sec:evaluation}

\subsection{Robot Dataset}
\label{sec:robot_dataset}
The dataset required to train and test the InEKFormer model was collected from simulation and real-world experiments conducted on the RH5 robotic platform, which is a 32-DoF humanoid robot equipped with various sensors, as depicted in Fig.~\ref{fig:rh5_frames_and_sensors}. The employed proprioceptive sensors operate at different frequencies as follows: IMU at 400~Hz, joint encoders for both linear and angular actuators at $\sim$150~Hz in average, and 6-axis force/torque (FT) sensors at the ankles at 1~kHz. The robot has closed-loop mechanisms in the ankles, knees, hips and torso. Thus, forward and inverse kinematics and dynamics for control and state estimation are computed using the HyRoDyn library~\cite{hyrodyn2020} for kinematics and dynamics of series-parallel hybrid mechanisms.

\includeimage{width=.56\linewidth}{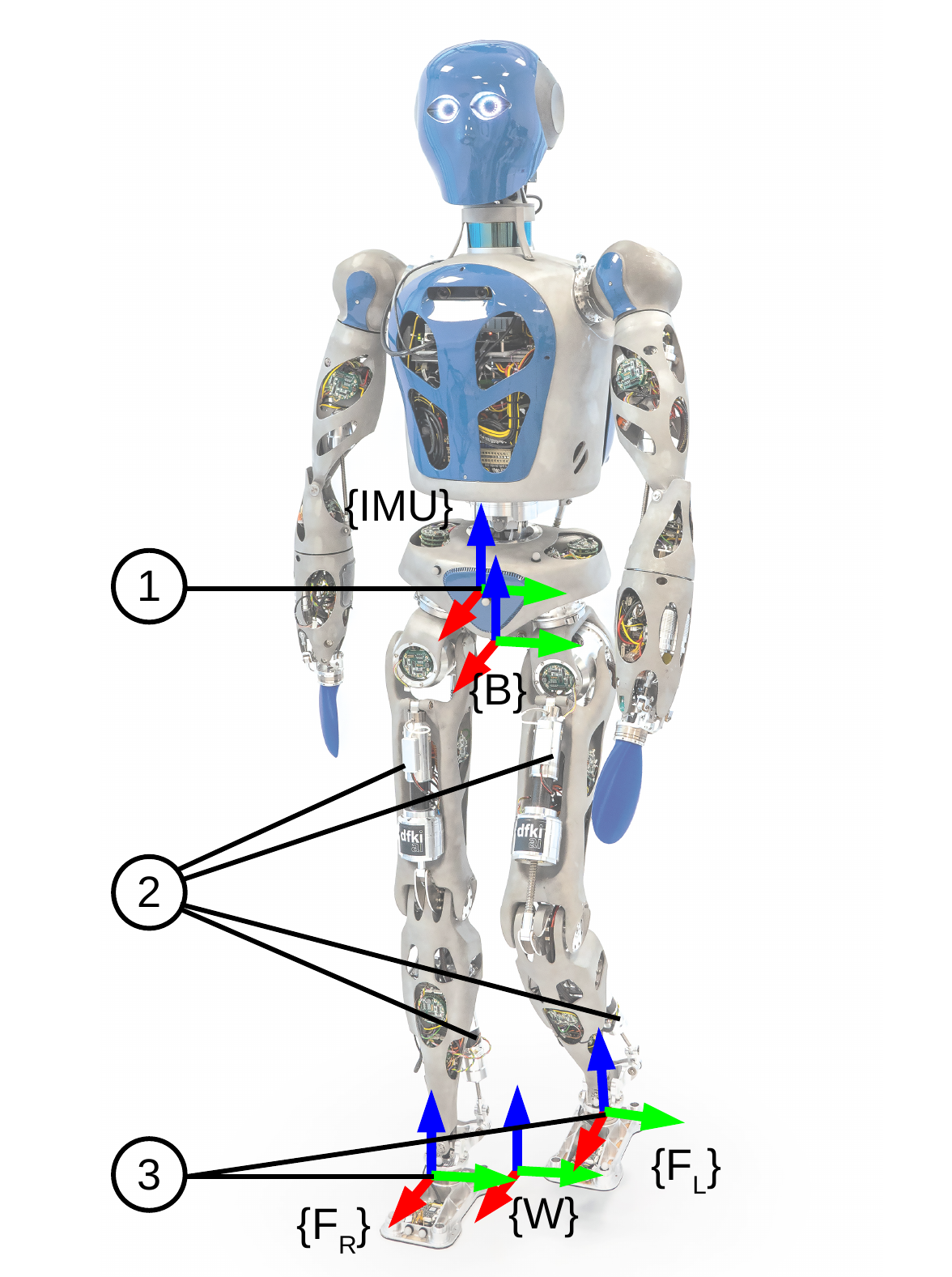}{RH5 robot coordinate frames and sensing hardware: 1 - IMU, 2 - Linear joint encoders, 3 - FT sensors.}{fig:rh5_frames_and_sensors}

The floating base frame $\{B\}$ provided by the state estimator is located at the waist and has the same orientation as the $\{IMU\}$ frame, but a different position. The foot frames $\{F_L\}$ and $\{F_R\}$ are located at the ankle joints, and the robot world frame $\{W\}$ is the projection of the floating base onto the $xy$-plane at the initial time $t_0$.

For data collection, the RH5 robot was controlled to perform various motions, such as static walking, squatting, balancing on a single leg, and waist movements along the sagittal and frontal axes.
The dataset, as shown in Fig.~\ref{fig:dataset}, consists of 30 different trajectories executed on the real system, as well as 14 additional trajectories collected in the RobotDART~\cite{RobotDART2024} simulation at 300 Hz. The recorded data includes ground truth of the floating base and feet poses, as well as IMU and joint encoder readings. 
The ground truth data has been recorded in the real-world experiments using a Qualisys motion capture setup with three cameras. Markers were placed on the robot's waist and feet and tracked at 300 Hz. After initial camera calibration, the maximum position residual on each Cartesian axis was 0.3~mm. Moreover, to generate a smooth ground truth and address short-term tracking inconsistencies, a Butterworth filter of order 3 has been applied. The real robot dataset was resampled at the frequency of the joint encoders, while the simulation dataset was downsampled at 150 Hz. Overall, a total of \num{575258} data samples were recorded over a time span of 50 minutes. The dataset is made openly available at \cite{Hohmeyer25}.

\includeimage{width=\linewidth}{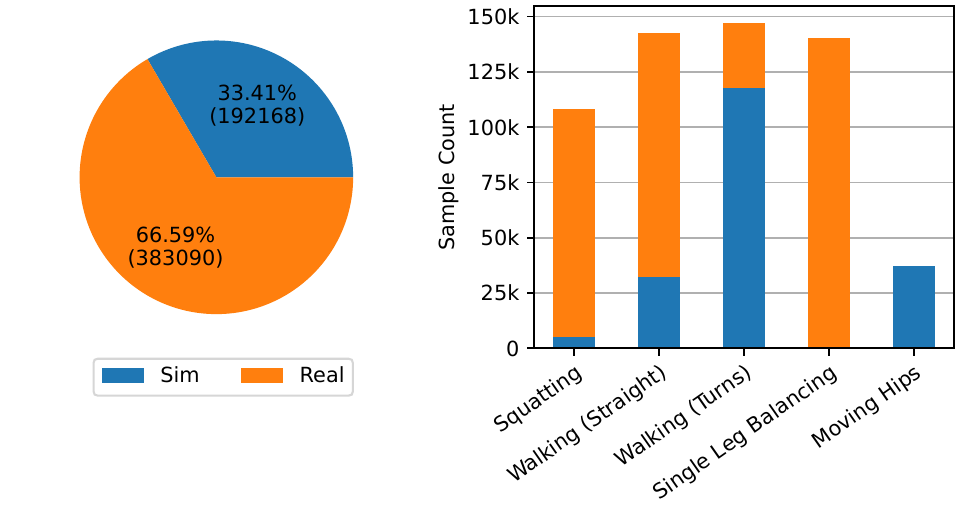}{Overview of the RH5 dataset.}{fig:dataset}

\subsection{Optimization and Training}

The gain estimator model is trained in a supervised fashion using a mean squared error (MSE) loss to compute the partial derivatives of the model parameters $\boldsymbol\Theta$. However, since the gain estimator model outputs a gain matrix, the corresponding corrected state $\mathbf{\bar{X}}^+$ must be calculated using (\ref{eq:correction}) to compute the distance to the ground truth state $\mathbf{X}$:
\begin{align}
\mathcal{L}(\boldsymbol\Theta) &= ||\mathbf{\bar{X}}^+_{}(\boldsymbol\Theta) - \mathbf{X}||^{2}_{\mathrm{F}} \nonumber\\
                    &= ||\exp(\mathbf{\hat{K}(\boldsymbol\Theta)\Pi\bar{X}Y})\mathbf{\bar{X}} - \mathbf{X}||^2_{\mathrm{F}}
\end{align}
\noindent where $||\cdot||_{\mathrm{F}}$ denotes the Frobenius norm of the corresponding matrix. The resulting gradients were backpropagated via the truncated backpropagation through time with scheduled sampling (TBPTT-SS) algorithm, which employs a logistic curve to transition smoothly over time from teacher-forcing to fully autoregressive training mode \cite{bengio2015}. Furthermore, an Adam optimizer together with OneCycleLR learning rate scheduler \cite{smith2019} was used. To optimize the different hyperparameters for the model and the training loop, a Tree-structured Parzen Estimator (TPE) sampler selects a set of hyperparameters for each training run to optimize the validation loss.

Although lower training and validation losses in a training run are strong indicators of a better gain estimator model, the final selection of the best-performing InEKFormer model was based on estimation results within the integrated filter loop. To compute the filter's accuracy, we define the absolute state estimation error per sample, denoted by $\boldsymbol\eta$, as follows:
\begin{equation}
	\boldsymbol{\eta} = \mathbf{\bar{X}}^+_{} - \mathbf{X}
\end{equation}

\noindent Moreover, the root mean-square error (RMSE) of an estimated trajectory of length N is obtained as follows:
\begin{equation}
	\boldsymbol{\eta}_{\mathrm{RMSE}} = \sqrt{\frac{1}{\mathrm{N}} \sum^{\mathrm{N}-1}_{i=0}\boldsymbol\eta_i^{2}}
\end{equation}

\subsection{Results}
To evaluate the InEKFormer method, different training and optimization setups were compared. The experiments are divided into three categories: (i) baseline tests, (ii) single trajectories and (iii) multiple trajectories. Table~\ref{tab:training} provides an overview of the trained models and the conditions for hyperparameter optimization. Additionally, Table~\ref{tab:hp_result} lists a subset of optimized hyperparameters for all trained models. The achieved RMSE values for the estimated trajectories of all models are listed in Table~\ref{tab:rmse}. 

\maketable{
	\resizebox{\columnwidth}{!}{
		\begin{tabular}{llrllrl}
			\toprule
			\multirow{2}{*}{Model} & \multicolumn{3}{l}{Train} & \multicolumn{3}{l}{Test}\\
			\cmidrule{2-7}
			& Data & Samples & Mode & Data & Samples & Mode \\
			\midrule
			$\boldsymbol\Omega_1$, $\boldsymbol\Gamma_1$ & Sim & \num{10} & TF & Sim & \num{20} & AR\\
			$\boldsymbol\Omega_2$ & Sim & \num{10} & AR & Sim & \num{20} & AR\\
			\midrule
			$\boldsymbol\Omega_3$, $\boldsymbol\Gamma_3$ & Real & \num{10} & TF & Real & \num{20} & AR\\
			$\boldsymbol\Omega_4$, $\boldsymbol\Gamma_4$ & Real & \num{10} & AR & Real & \num{20} & AR\\
			\midrule
			$\boldsymbol\Omega_5$, $\boldsymbol\Gamma_5$ & Sim & \num{7835} & TF & Real & \num{9899} & 1A\\
			$\boldsymbol\Omega_6$, $\boldsymbol\Gamma_6$ & Real & \num{7425} & TF & Real & \num{9899} & 1A\\
			\midrule
			$\boldsymbol\Omega_7$ & Sim & \num{85296} & TF & Real & \num{9899} & 1A\\
			$\boldsymbol\Omega_8$ & Real & \num{373191} & TF & Real & \num{9899} & 1A\\
			$\boldsymbol\Omega_9$ & Sim + Real & \num{565359} & TF & Real & \num{9899} & 1A\\
			\bottomrule
		\end{tabular}
}}{Gain estimator models and training conditions: $\boldsymbol\Omega$ - InEKFormer, $\boldsymbol\Gamma$ - KalmanNet, TF - teacher-forcing, AR - autoregressive, 1A - one-step-ahead mode.}{tab:training}

\maketable{
	
		\begin{tabular}{lrlr}
			\toprule
			Model & Parameters & Activations & $n_{\mathrm{history}}$\\
			\midrule
			$\boldsymbol\Omega_1$ & \num{61474} & GELU, Tanh & \num{10}\\
			$\boldsymbol\Omega_2$ & \num{56410} & ReLU, Tanh & \num{10}\\
			$\boldsymbol\Omega_3$ & \num{56410} & ReLU, Tanh & \num{10}\\
			$\boldsymbol\Omega_4$ & \num{54950} & ReLU, Tanh & \num{10}\\
			$\boldsymbol\Omega_5$ & \num{214298} & GELU & \num{10}\\
			$\boldsymbol\Omega_6$ & \num{97562} & GELU & \num{50}\\
			$\boldsymbol\Omega_7$ & \num{517094} & GELU, Tanh & \num{100}\\
			$\boldsymbol\Omega_8$ & \num{202114} & GELU & \num{100}\\
			$\boldsymbol\Omega_9$ & \num{202114} & GELU & \num{100}\\
            \midrule
			$\boldsymbol\Gamma_1$ & \num{740829} & ReLU & $\infty$\\
			$\boldsymbol\Gamma_3$ & \num{740829} & ReLU & $\infty$\\
			$\boldsymbol\Gamma_4$ & \num{740829} & ReLU & $\infty$\\
			$\boldsymbol\Gamma_5$ & \num{1013301} & ReLU & $\infty$\\
			$\boldsymbol\Gamma_6$ & \num{877065} & ReLU & $\infty$\\
			\bottomrule
		\end{tabular}
}{Hyperparameter subsets for InEKFormer ($\boldsymbol\Omega$ models) and KalmanNet ($\boldsymbol\Gamma$ models).}{tab:hp_result}

\maketabletwocolumns{
	\begin{tabular}{llllllllllll}
		\toprule
		Test & Filter & Mode & $\phi_{\text{x}}\ [\mathrm{rad}]$ & $\phi_{\text{y}}\ [\mathrm{rad}]$ & $\phi_{\text{z}}\ [\mathrm{rad}]$ & $\mathrm{v}_{\text{x}}\ [\frac{\mathrm{m}}{\mathrm{s}}]$ & $\mathrm{v}_{\text{y}}\ [\frac{\mathrm{m}}{\mathrm{s}}]$ & $\mathrm{v}_{\text{z}}\ [\frac{\mathrm{m}}{\mathrm{s}}]$ & $\mathrm{p}_{\text{x}}\ [\mathrm{m}]$ & $\mathrm{p}_{\text{y}}\ [\mathrm{m}]$ & $\mathrm{p}_{\text{z}}\ [\mathrm{m}]$\\
		\midrule
		\multirow{7}{*}{i} & $\boldsymbol{\Omega}_1$ & AR & $1.298$ & $0.217$ & $0.844$ & $4.482$ & $3.014$ & $2.494$ & $0.695$ & $1.052$ & $2.381$\\
		& $\boldsymbol{\Omega}_2$ & AR & $\mathbf{0.036}$ & $\mathbf{0.053}$ & $\mathbf{0.051}$ & $\mathbf{0.071}$ & $\mathbf{0.084}$ & $\mathbf{0.384}$ & $\mathbf{0.063}$ & $\mathbf{0.017}$ & $\mathbf{0.063}$\\
		& $\boldsymbol{\Gamma}_1$ & AR & $1.048$ & $1.568$ & $3.738$ & $0.248$ & $0.277$ & $2.98e3$ & $0.417$ & $0.332$ & $9.528$\\
		\cmidrule{2-12}
		& $\boldsymbol{\Omega}_3$ & AR & $0.853$ & $0.206$ & $0.742$ & $0.222$ & $0.199$ & $0.442$ & $1.085$ & $0.503$ & $0.397$\\
		& $\boldsymbol{\Omega}_4$ & AR & $\mathbf{5e\text{-}4}$ & $\mathbf{0.004}$ & $\mathbf{9e\text{-}4}$ & $\mathbf{0.012}$ & $\mathbf{0.002}$ & $\mathbf{0.011}$ & $\mathbf{0.002}$ & $\mathbf{0.003}$ & $\mathbf{0.009}$\\
		& $\boldsymbol{\Gamma}_3$ & AR & $0.170$ & $0.070$ & $0.044$ & $0.054$ & $0.039$ & $0.088$ & $0.111$ & $0.342$ & $0.882$\\
		& $\boldsymbol{\Gamma}_4$ & AR & $0.215$ & $0.063$ & $0.052$ & $0.080$ & $0.092$ & $0.136$ & $0.134$ & $0.339$ & $0.921$\\
		\midrule
		\multirow{4}{*}{ii} & $\boldsymbol{\Omega}_5$ & 1A & $\mathbf{0.029}$ & $0.032$ & $\mathbf{0.023}$ & $0.055$ & $0.042$ & $2.484$ & $0.035$ & $0.042$ & $0.052$\\
		& $\boldsymbol{\Omega}_6$ & 1A & $0.042$ & $\mathbf{0.003}$ & $0.027$ & $\mathbf{0.034}$ & $\mathbf{0.012}$ & $\mathbf{0.043}$ & $\mathbf{0.017}$ & $\mathbf{0.027}$ & $\mathbf{0.009}$\\
		& $\boldsymbol{\Gamma}_5$ & 1A & $0.062$ & $0.079$ & $0.107$ & $0.067$ & $0.060$ & $43.62$ & $0.287$ & $0.306$ & $0.717$\\
		& $\boldsymbol{\Gamma}_6$ & 1A & $0.061$ & $0.081$ & $0.114$ & $0.059$ & $0.068$ & $0.139$ & $0.287$ & $0.307$ & $0.840$\\
		\midrule
		\multirow{4}{*}{iii} & $\boldsymbol\Omega_7$ & 1A & $0.601$ & $0.406$ & $1.132$ & $11.78$ & $14.06$ & $40.20$ & $0.363$ & $1.633$ & $0.686$\\
		& $\boldsymbol\Omega_8$ & 1A & $0.016$ & $0.006$ & $0.003$ & $0.004$ & $0.007$ & $0.153$ & $0.005$ & $0.016$ & $0.003$\\
		& $\boldsymbol\Omega_9$ & 1A & $\mathbf{8e\text{-}4}$ & $\mathbf{0.001}$ & $\mathbf{8e\text{-}4}$ & $\mathbf{0.003}$ & $\mathbf{0.005}$ & $0.187$ & $\mathbf{7e\text{-}4}$ & $\mathbf{0.001}$ & $\mathbf{6e\text{-}4}$\\
		& InEKF & AR & $0.043$ & $0.008$ & $0.006$ & $0.052$ & $0.071$ & $\mathbf{0.027}$ & $0.020$ & $0.038$ & $0.003$\\
		\bottomrule
	\end{tabular}
}{RMSE per state dimension of the InEKFormer ($\boldsymbol{\Omega}_1$ to $\boldsymbol{\Omega}_{9}$), the KalmanNet ($\boldsymbol{\Gamma}_{1}$, $\boldsymbol{\Gamma}_{3}$ to $\boldsymbol{\Gamma}_{6}$) and the online InEKF. (AR - autoregressive mode, 1A - one-timestep-ahead mode).}{tab:rmse}

\subsubsection*{i) Baseline Tests} Firstly, baseline overfitting tests on short sequences were conducted to ensure that the proposed {{InEKFormer}} algorithm works as intended and the RH5 dataset provides sufficient information. The gain estimator models $\boldsymbol{\Omega}_1$ to $\boldsymbol{\Omega}_4$ were trained on the first 10 samples of a simulated and a real RH5 walking motion, respectively, and tested on the first 20 samples of the same motions. Training was done either in autoregressive or teacher-forcing mode. For the sake of comparison, the KalmanNet models $\boldsymbol\Gamma_1$, $\boldsymbol\Gamma_3$ and $\boldsymbol\Gamma_4$ were trained under the same conditions. $\boldsymbol\Gamma_2$ is missing since the autoregressive training on simulation data did not converge to a satisfactory loss value.

The absolute position estimation errors of the models that were trained on simulation data ($\boldsymbol{\Omega}_1$, $\boldsymbol{\Omega}_2$, $\boldsymbol{\Gamma}_1$) are shown in Fig.~\ref{fig:error_omega1_4}. The results show that the InEKFormer is generally effective, as the models were able to overfit to the first 10 samples of each sequence, while the estimations for the subsequent 10 non-training samples gradually deviate from the real trajectory. This demonstrates that the training data is not simply reproduced, but that general patterns can be recognized, some of which extend to unseen data. However, $\boldsymbol\Gamma_1$ is oscillating around the first 10  samples, indicating that the KalmanNet approach, based on RNNs, is not well suited to the high-dimensional data of the RH5 humanoid robot.

Moreover, as shown in Table~\ref{tab:rmse}, the autoregressive training mode is superior to the teacher forcing mode, as $\boldsymbol\Omega_2$, $\boldsymbol\Omega_4$ and $\boldsymbol\Gamma_4$ perform up to $10^{3}$ better than $\boldsymbol\Omega_1$, $\boldsymbol\Omega_3$ and $\boldsymbol\Gamma_3$. Additionally, the estimation error of the KalmanNet models are magnitudes greater than those of the InEKFormer models, thus proving the superiority of the InEKFormer. Finally, this test shows that the dataset and feature selection provide the necessary causal information for Kalman gain prediction.

\includeimage{width=\linewidth}{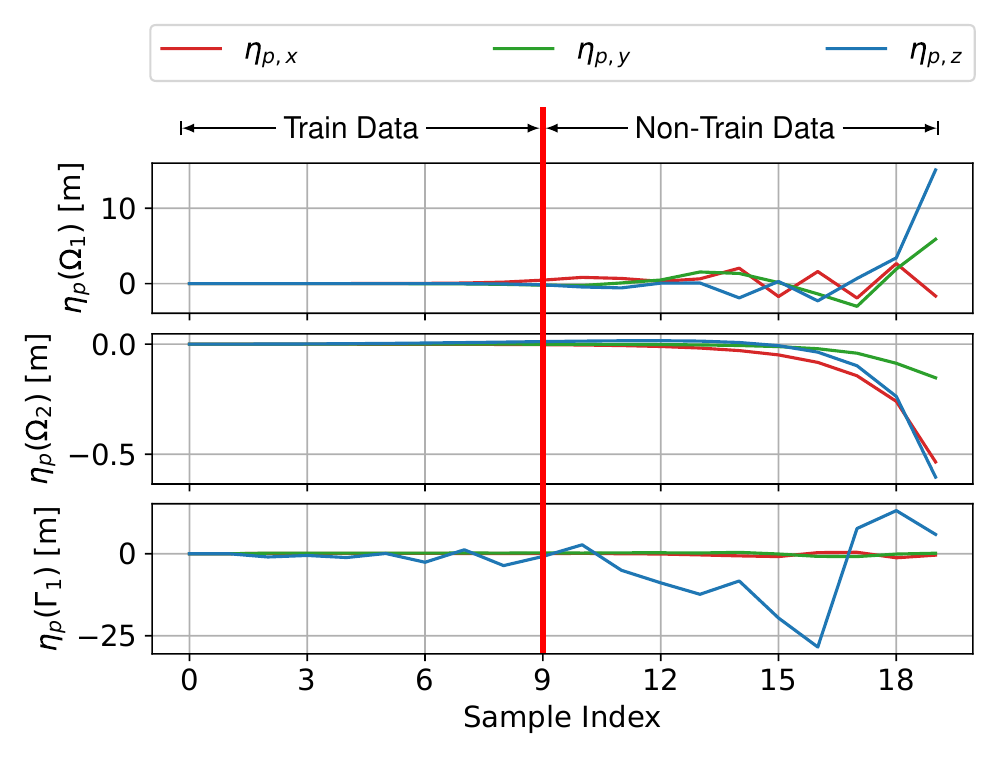}{Absolute position estimation errors $\boldsymbol\eta_{\mathrm{p}}$ on the first 20 samples from the simulated walking test motion for models $\boldsymbol{\Omega}_1$, $\boldsymbol{\Omega}_2$ and $\boldsymbol{\Gamma}_1$.}{fig:error_omega1_4}

\subsubsection*{ii) Single Trajectory Tests} Benchmarking tests were conducted for single trajectories to compare the InEKFormer with KalmanNet and the model-based InEKF on a real RH5 walking motion. For this purpose, two Transfomer-based gain estimator models, $\boldsymbol{\Omega}_5$ and $\boldsymbol{\Omega}_6$, as well as two KalmanNet models, $\boldsymbol{\Gamma}_5$ and $\boldsymbol{\Gamma}_6$, were trained on the first 75\% of a single simulated and real-world walking trajectory, respectively, while the rest of the motion was used for testing. As autoregressive training was not successful, the models were trained in teacher-forcing mode instead. This experiment investigates the models' capability to generalize to unseen data and to transfer patterns from simulation to real robot data. The corresponding estimated trajectories of the test motion are shown in Fig.~\ref{fig:t3_trajectories}. For visibility reasons, model $\boldsymbol\Omega_5$ is not displayed, since it exhibits high-frequency transverse oscillations around the ground truth trajectory.

The RMSE values in Table~\ref{tab:rmse} show that $\boldsymbol\Omega_6$ outperforms $\boldsymbol\Omega_5$ in all state dimensions except $x$ and $z$-axis rotation. Moreover, both gain estimator models are outperformed by the InEKF in terms of orientation, velocity, and position in the $z$-axis. Fig.~\ref{fig:t3_trajectories} shows that, even when trained in teacher-forcing mode, the KalmanNet models $\boldsymbol{\Gamma}_5$ and $\boldsymbol{\Gamma}_6$ were unable to learn the patterns of the robot motions, proving the limitations of KalmanNet on high-dimensional data.

\subsubsection*{iii) Multiple Trajectory Tests} Finally, the models $\boldsymbol{\Omega}_7$, $\boldsymbol{\Omega}_{8}$ and $\boldsymbol{\Omega}_{9}$ were trained on datasets containing multiple motion trajectories and tested on the RH5 walking motion that has been used in the single trajectory test. Model $\boldsymbol\Omega_7$ was trained on simulation data only, and $\boldsymbol\Omega_8$ was trained on real robot data, excluding the test trajectory. Model $\boldsymbol\Omega_9$ was trained on the entire dataset except the test trajectory.  Note that training of the KalmanNet failed for all these models, as its loss function did not converge during training. The estimated trajectories of the floating base are shown in Fig.~\ref{fig:t3_trajectories}.

The results show that, in general, more training data leads to better filter performance. This is evident in Table~\ref{tab:rmse}, which shows that RMSE values decrease as the number of training samples increases. The only exception to this pattern is $\boldsymbol\Omega_7$, which has been excluded from Fig.~\ref{fig:t3_trajectories} for clarity. It exhibits heavier transversal oscillations around the ground truth trajectory than $\boldsymbol\Omega_5$, possibly due to the sim-to-real gap. 

\includeimage{width=\linewidth}{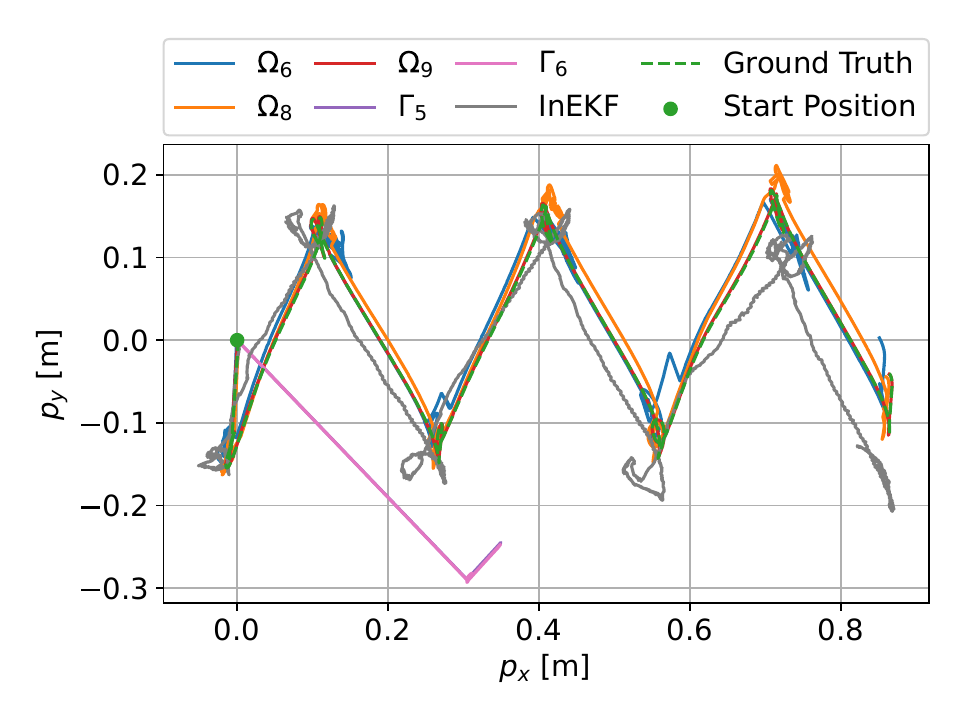}{One-timestep-ahead estimates of the $x$ and $y$ positions of the models from single ($\boldsymbol{\Omega}_6$, $\boldsymbol{\Gamma}_5$, $\boldsymbol{\Gamma}_6$) and multiple trajectory tests ($\boldsymbol{\Omega}_8$, $\boldsymbol{\Omega}_9$), as well as online InEKF on a walking trajectory.}{fig:t3_trajectories} 

As shown in Table~\ref{tab:hp_result}, all KalmanNet models have up to 10 times more trainable parameters than the corresponding InEKFormer gain estimator models despite lower estimation performance. This clearly demonstrates InEKFormer's superiority over KalmanNet in terms of model size efficiency when dealing with high-dimensional data. Additionally, most models trained on simulated data only ($\boldsymbol\Omega_1$, $\boldsymbol\Omega_2$, $\boldsymbol\Omega_5$, $\boldsymbol\Omega_7$, $\boldsymbol\Gamma_5$) required more trainable parameters than models trained on real robot data, due to the dataset inherent sim-to-real gap.

\subsection{Discussion}

Since a stable autoregressive training for single and multiple motion trajectories was not possible with standard regularization methods (e.g., gradient clipping, small learning rates, a generic loss function and scheduled sampling), the respective models were optimized in teacher-forcing mode. Testing was done in a closed filter loop with ground truth feedback (aka. one-timestep-ahead predictions) since the exposure bias resulting from the training in teacher-forcing mode would have grown excessively with more than one-timestep-ahead predictions. Clearly, Transformer models are much more challenging to train for regression tasks than for classification tasks, such as natural language processing (NLP). Regression tasks inherently require exactly one correct output per training sample. In contrast, there are theoretically infinite correct outputs for classifying a single sample, as long as the probability of the correct target class is the highest. Possible solutions might include more sophisticated regularization techniques, such as curriculum learning, or enhanced sequence preprocessing.

Additionally, the results in Table~\ref{tab:rmse} indicate the potential for sim-to-real transfer learning, especially for a single motion. For example, the performance of the model $\boldsymbol\Omega_5$ trained on simulation data alone is better than model $\boldsymbol\Omega_6$ trained on real data for the orientation in the $x$ and $z$-direction. However, with multiple motions, the difference between the simulated and real robot increases, as indicated by the fact that the RMSE values of the models trained on real data, namely $\boldsymbol\Omega_8$ and $\boldsymbol\Omega_9$, are smaller by a factor of 100 compared to $\boldsymbol\Omega_7$, which was trained on simulation data. Moreover, Table~\ref{tab:rmse} also shows that the InEKF estimates are best in $z$-direction. This suggests that the dataset needs improvement, particularly in this dimension. Even the best-performing model $\boldsymbol\Omega_9$ was outperformed by the InEKF in the $z$-direction velocity, despite the InEKF's constant drift, as shown in Fig.~\ref{fig:t3_trajectories}. Hence, a more realistic simulation with an improved robot model could reduce the sim-to-real gap.

Furthermore, the transversal oscillations in the estimated trajectories of $\boldsymbol\Omega_5$ and $\boldsymbol\Omega_7$ can partly be explained by inconsistent sampling frequencies in the simulation and on the real robot. This is primarily caused by the varying sampling frequencies of the robot joint encoders, which operate at an average frequency of $\sim$150 Hz, ranging between 120-190 Hz on the real hardware. This leads to over-correcting behavior in the respective filters, resulting in transversal oscillations. This behaviour gets worse if the model was trained on multiple motions, such as $\boldsymbol\Omega_7$. To counteract this, the $\Delta t$ for each time step could be given as an input to the gain estimator model, which should incorporate a spatiotemporal attention mechanism to explicitly account for varying temporal token distances of different sensor modalities.

In terms of execution speed, the average inference time for the optimized models in the closed filter loop is between 50 - 350 ms executed on a 4-core i5-3320M CPU running at 2.60 GHz. 
These inference times can be considered as hardware-induced upper bounds. Modern robots can be equipped with fast, specialised processors that significantly reduce computation time. In addition, the InEKFormer algorithm could be improved by incorporating smaller and more efficient gain estimator models \cite{Yi2022} than the standard Transformer with scaled dot-product attention. This would lead to shorter inference times and enhanced real-time capability of the InEKFormer algorithm.

\section{CONCLUSION}
\label{sec:conclusion}

In this work, we presented the novel InEKFormer algorithm and applied it to the task of floating base state estimation on a simulated and real RH5 humanoid robot. We compared the approach to the model-based InEKF and the hybrid KalmanNet in terms of estimation accuracy. It was shown that autoregressive training of the Transformer gain estimator model is essential for the application of the InEKFormer in a practical scenario with estimation feedback. Although it was not possible to perform stable autoregressive training on larger datasets, it could be shown that the InEKFormer outperforms KalmanNet on short trajectories with full estimation feedback and that it outperforms both KalmanNet and InEKF on longer trajectories with ground truth feedback only. Furthermore, the potential for sim-to-real transfer learning was demonstrated.

As future work, the major shortcoming to be tackled is achieving stable autoregressive training, which would allow the model to be executed online on the robot. This could be realized using advanced regularization methods and more sophisticated Transformer architectures which leverage the invariance of Lie groups. Secondly, the sim-to-real gap can be addressed by conducting an ablation study to determine the optimal ratio of real to simulated robot data, as well as by improving the quality and quantity of the RH5 dataset. Finally, the proposed method could be generalized beyond humanoids, for instance to quadrupeds, aerial robots, or wheeled mobile robots. 

%%%%%%%%%%%%%%%%%%%%%%%%%%%%%%%%%%%%%%%%%%%%%%%%%%%%%%%%%%%%%%%%%%%%%%%%%%%%%%%%
%\section*{APPENDIX}

\section*{ACKNOWLEDGMENT}

This research was done in the CoEx (Grant number 01IW24008) and ActGPT (Grant number 01IW25002) projects funded by the German Aerospace Center (DLR) with federal funds from the German Ministry of Research, Technology and Space (BMFTR).

%%%%%%%%%%%%%%%%%%%%%%%%%%%%%%%%%%%%%%%%%%%%%%%%%%%%%%%%%%%%%%%%%%%%%%%%%%%%%%%%

\bibliographystyle{IEEEtranBST/IEEEtran}
\bibliography{IEEEtranBST/IEEEabrv,IEEEtranBST/main}

\end{document}